\documentclass[12pt]{article}
\usepackage[numbers,sort&compress]{natbib}
\usepackage[final]{neurips_2018}
\usepackage[utf8]{inputenc}
\usepackage[T1]{fontenc}
\usepackage{amsmath,amsfonts,amssymb,bm,bbm}
\usepackage{xifthen,xargs}
\usepackage{pifont}
\usepackage{nicefrac}
\usepackage{mathtools}
\usepackage{enumerate}
\usepackage{enumitem}
\usepackage{hyperref}
\usepackage[font=small,labelfont=bf]{caption}
\usepackage{wrapfig}
\usepackage{tabularx}
\usepackage{booktabs}
\usepackage{algorithm,algpseudocode}
\usepackage{bbding}
\usepackage{array}
\usepackage{relsize}

\usepackage[pdftex,dvipsnames]{xcolor}
\usepackage[colorinlistoftodos,prependcaption,textsize=tiny]{todonotes}

\newcommandx{\flag}[2][1=]
  {\todo[linecolor=red,backgroundcolor=red!25,bordercolor=red,#1]{#2}}
\newcommandx{\ask}[2][1=]
  {\todo[linecolor=OliveGreen,backgroundcolor=OliveGreen!25,bordercolor=OliveGreen,#1]{#2}}
\newcommandx{\note}[2][1=]
  {\todo[linecolor=blue,backgroundcolor=blue!25,bordercolor=blue,#1]{#2}}
 \newcommandx{\reply}[2][1=]
  {\todo[linecolor=Gray,backgroundcolor=Gray!25,bordercolor=Gray,#1]{#2}}

\newcommand{\temp}[1]{\textcolor{black}{#1}}
\newcommand{\warn}[1]{{\textcolor{black}{#1}}}

\newcommand{\marc}[1]{\textcolor{black}{#1}}

\newcommand{\eg}{\text{e.g.\@}}
\newcommand{\ie}{\text{i.e.\@}}

\newcommand{\wrt}{\text{w.r.t.\@}}

\newcommand{\rvs}{\text{rvs}}

\newcommand{\subheading}[1]{\paragraph{#1}}

\newcommand{\inlineItem}[1]{#1)}

\newcommand{\blackbox}{black-box}
\newcommand{\mmform}{incremental}

\newcommand{\legendDot}{\raisebox{1.25pt}{$\scriptscriptstyle{\bullet}$}}
\newcommand{\legendStar}{\raisebox{-1.75pt}{\FiveStarOpen}}
\newcommand{\legendX}{\ensuremath{\bm{\times}}}

\newcommand{\E}{\mathbb{E}}
\newcommand{\T}[1]{\ensuremath{#1^\top}}

\newcommand{\opt}[1]{\ensuremath{#1^{*}}}
\newcommand{\alt}[1]{\ensuremath{#1^\prime}}

\newcommand{\mathhyphen}{{\hbox{-}}}
\newcommand{\qeq}{\ensuremath{\stackrel{?}{=}}}

\newcommand{\reals}{\mathbb{R}}
\newcommand{\data}{\ensuremath{\mathcal{D}}}
\newcommand{\model}{\ensuremath{\mathcal{M}}}
\newcommand{\belief}{\ensuremath{p}}

\newcommand{\nSamples}{\ensuremath{m}}
\newcommand{\nEvals}{\ensuremath{N}}

\newcommand{\sampleIdx}{\ensuremath{k}}
\newcommand{\sample}[1]{\ensuremath{#1^{\sampleIdx}}}

\newcommand{\roundIdx}{\ensuremath{j}}
\newcommand{\iAcq}{\ensuremath{\bar{\Acq}}}

\newcommand{\oldSubscript}{\ensuremath{\mathsmaller{<} \roundIdx}}

\newcommand{\NEW}[2]{\ensuremath{{\ifthenelse{\equal{#1}{1}}{#2_{\roundIdx}}{#2_{\roundIdx}}}}}
\newcommand{\OLD}[2]{\ensuremath{{\ifthenelse{\equal{#1}{1}}{#2_{\oldSubscript}}{#2_{\oldSubscript}}}}}

\newcommand{\BOTH}[2]{\ensuremath{{\ifthenelse{\equal{#1}{1}}{#2_{1:\roundIdx}}{#2_{1:\roundIdx}}}}}

\newcommand{\solver}{maximizer}
\newcommand{\Solver}{Maximizer}
\newcommand{\solvers}{\solver s}
\newcommand{\Solvers}{\Solver s}

\newcommand{\parallelism}{\ensuremath{q}}

\newcommand{\x}{\ensuremath{\mathbf{x}}}
\newcommand{\X}{\ensuremath{\mathbf{X}}}
\newcommand{\xspace}{\ensuremath{\mathcal{X}}}
\newcommand{\xdim}{\ensuremath{d}}

\newcommand{\y}{\ensuremath{y}}
\newcommand{\Y}{\ensuremath{\mathbf{y}}}
\newcommand{\yspace}{\ensuremath{\mathcal{Y}}}

\newcommand{\z}{\ensuremath{z}}
\newcommand{\Z}{\ensuremath{\mathbf{z}}}
\newcommand{\zspace}{\ensuremath{\mathcal{Z}}}

\newcommand{\event}{\ensuremath{\boldsymbol{e}}}
\newcommand{\events}{\ensuremath{\boldsymbol{e}}}
\newcommand{\espace}{\ensuremath{\mathcal{E}}}

\newcommand{\f}{\ensuremath{f}}

\newcommand{\gaussian}{\ensuremath{\mathcal{N}}}
\newcommand{\means}{\ensuremath{\boldsymbol{\mu}}}
\newcommand{\mean}{\ensuremath{\mu}}

\newcommand{\Cov}{\ensuremath{\boldsymbol{\Sigma}}}

\newcommand{\Precis}{\ensuremath{\boldsymbol{\Lambda}}}

\newcommand{\chol}{\ensuremath{L}}
\newcommand{\Chol}{\ensuremath{\mathbf{L}}}
\newcommand{\stddev}{\ensuremath{\sigma}}

\newcommand{\var}{\ensuremath{\stddev^{2}}}

\newcommand{\resid}{\ensuremath{\gamma}}
\newcommand{\resids}{\ensuremath{\boldsymbol{\gamma}}}

\newcommand{\level}{\ensuremath{\alpha}}

\newcommand{\reparam}{\ensuremath{\phi}}
\newcommand{\rparam}{\ensuremath{\theta}}

\newcommand{\rparams}{\ensuremath{\boldsymbol{\theta}}}
\newcommand{\rParams}{\ensuremath{\boldsymbol{\Theta}}}

\newcommand{\hypers}{\ensuremath{\boldsymbol{\ensuremath{\zeta}}}}

\newcommand{\zeros}{\ensuremath{\mathbf{0}}}
\newcommand{\identity}{\ensuremath{\mathbf{I}}}

\newcommand{\half}{\ensuremath{\nicefrac{1}{2}}}

\newcommand{\lbounds}{\ensuremath{\boldsymbol{a}}}

\newcommand{\ubounds}{\ensuremath{\boldsymbol{b}}}

\newcommand{\relu}{\ensuremath{\operatorname{ReLU}}}
\newcommand{\sigmoid}{\ensuremath{\operatorname{\sigma}}}
\newcommand{\softmax}{\ensuremath{\operatorname{softmax}}}
\newcommand{\heaviside}{\ensuremath{\operatorname{\mathbbm{1}}}}

\newcommand{\entropy}{\ensuremath{\operatorname{H}}}

\newcommand{\acq}{\ensuremath{\ell}}
\newcommand{\Acq}{\ensuremath{\mathcal{L}}}
\newcommand{\AcqMC}[1]{\ensuremath{\Acq_{\ifthenelse{\isempty{#1}}{\nSamples}{#1}}}}

\newcommand{\acqParams}{\ensuremath{\boldsymbol{\psi}}}

\newcommand{\temperature}{\ensuremath{\tau}}
\newcommand{\parallelize}[1]{\ensuremath{\operatorname{\mathnormal{q}\mathhyphen{}#1}}}
\newcommand{\acqSubscript}[1]{\ensuremath{\scriptscriptstyle{#1}}}

\newcommand{\EI}{\ensuremath{\operatorname{EI}}}
\newcommand{\UCB}{\ensuremath{\operatorname{UCB}}}
\newcommand{\PI}{\ensuremath{\operatorname{PI}}}
\newcommand{\SR}{\ensuremath{\operatorname{SR}}}
\newcommand{\ES}{\ensuremath{\operatorname{ES}}}
\newcommand{\KG}{\ensuremath{\operatorname{KG}}}

\newcommand{\qEI}{\parallelize{EI}}
\newcommand{\qUCB}{\parallelize{UCB}}

\newcommand{\MC}{MC{} }

\newcommand{\setA}{\ensuremath{\mathcal{A}}}
\newcommand{\setB}{\ensuremath{\mathcal{B}}}
\newcommand{\setC}{\ensuremath{\mathcal{C}}}
\newcommand{\lboundAcq}{\ensuremath{v_{\min}}}

\newcommand{\matern}[2]{Mat\'{e}rn-{\nicefrac{#1}{#2}}}

\newcommand{\adam}{\textsc{Adam}}
\newcommand{\direct}{\text{DIRECT}}

\newcommand{\abs}[1]{\ensuremath{\lvert#1\rvert}}

\DeclarePairedDelimiter{\floor}{\lfloor}{\floor}

\newcommand{\argmax}{\ensuremath{\operatornamewithlimits{arg\,max}}}

\newcommand{\gain}{\ensuremath{\delta}}
\newcommand{\Gain}{\ensuremath{\Delta}}
\newcommand{\argdot}{\makebox[1ex]{\textbf{$\cdot$}}}
\newcommand{\MM}{MM}
\newcommand{\SM}{SM}
\newcommand{\uacq}{\ensuremath{\hat{\acq}}}

\newcommand{\vid}{k}
\newcommand{\version}[1]{\ensuremath{#1^{\vid}}}

\begin{document}


\title{Maximizing acquisition functions\\ for Bayesian optimization}
\author{
 	James T. Wilson\thanks{Correspondence to j.wilson17@imperial.ac.uk} \\
    {Imperial College London}
 	\And
 	Frank Hutter\\
    {University of Freiburg}
 	\And
 	Marc Peter Deisenroth\\
    {Imperial College London}\\
    {PROWLER.io}
}
\maketitle

\begin{abstract}
Bayesian optimization is a sample-efficient approach to global optimization that relies on theoretically motivated value heuristics (acquisition functions) to guide its search process. Fully maximizing acquisition functions produces the Bayes' decision rule, but this ideal is difficult to achieve since these functions are frequently non-trivial to optimize. This statement is especially true when evaluating queries in parallel, where acquisition functions are routinely non-convex, high-dimensional, and intractable.
We first show that acquisition functions estimated via Monte Carlo integration are consistently amenable to gradient-based optimization. Subsequently, we identify a common family of acquisition functions, including \EI{} and \UCB{}, whose properties not only facilitate but justify use of greedy approaches for their maximization.

%
\end{abstract}


\section{Introduction}
\label{sect:introduction}

Bayesian optimization (BO) is a powerful framework for tackling complicated global optimization problems \cite{kushner1964new,movckus1975bayesian,jones-jgo98a}. Given a \blackbox{} function \(\f : \xspace \to \yspace\), BO seeks to identify a maximizer \(\opt{\x} \in \argmax_{\x \in \xspace} \f(\x)\) while simultaneously minimizing incurred costs. Recently, these strategies have demonstrated state-of-the-art results on many important, real-world problems ranging from material sciences \cite{frazier2016bayesian,ueno2016combo}, to robotics \cite{calandra2016bayesian,bansal2017goal}, to algorithm tuning and configuration \cite{hutter2011sequential,snoek-nips12a,swersky-nips13,falkner-icml-18}.

From a high-level perspective, BO can be understood as the application of Bayesian decision theory to optimization problems~\cite{movckus1994application, degroot2005optimal,robert2007bayesian}. One first specifies a belief over possible explanations for \f{} using a probabilistic surrogate model and then combines this belief with an acquisition function \Acq{} to convey the expected utility for evaluating a set of queries \X{}. 
In theory, \X{} is chosen according to Bayes' decision rule as \Acq{}'s maximizer by solving for an \emph{inner optimization problem} \cite{gelbart2014bayesian,martinez2014bayesopt,wang2016parallel}. 
In practice, challenges associated with maximizing \Acq{} greatly impede our ability to live up to this standard.
Nevertheless, this inner optimization problem is often treated as a \blackbox{} unto itself. 
Failing to address this challenge leads to a systematic departure from BO's premise and, consequently, consistent deterioration in \temp{achieved} performance.

To help reconcile theory and practice, we present two modern perspectives for addressing BO's inner optimization problem that exploit key aspects of acquisition functions and their estimators.
First, we clarify how sample path derivatives can be used to optimize a wide range of acquisition functions estimated via Monte Carlo (MC) integration.
Second, we identify a common family of submodular acquisition functions and show that its constituents can generally be expressed in a \temp{more computer-friendly} form. These acquisition functions' properties enable greedy approaches to efficiently maximize them with guaranteed near-optimal results.
Finally, we demonstrate through comprehensive experiments that these theoretical contributions directly translate to reliable and, often, substantial performance gains. 


\section{Background}
\label{sect:background}
\begin{figure}[t]
\begin{center}
\includegraphics[width=\linewidth]{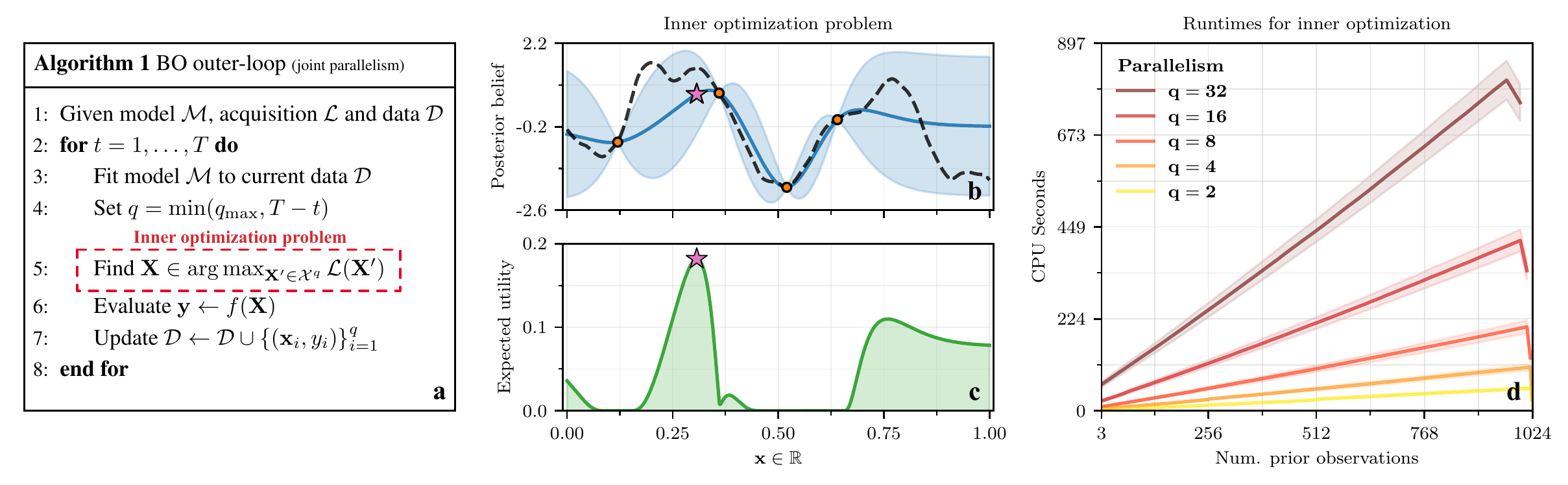}
\vspace{-20pt}
\caption{(a) Pseudo-code for standard BO's ``outer-loop'' with parallelism \parallelism{}; the inner optimization problem is boxed in red. (b--c) GP-based belief and expected utility (EI), given four initial observations `\legendDot'. The aim of the inner optimization problem is to find the optimal query `\legendStar'. (d) Time to compute \(2^{14}\) evaluations of \MC \qEI{} using a GP surrogate for varied observation counts and degrees of parallelism. Runtimes fall off at the final step because \parallelism{} decreases to accommodate evaluation budget \(T=1,024\).
}
\label{fig:overview_pt1}
\end{center}
\end{figure}

Bayesian optimization relies on both a surrogate model \model{} and an acquisition function \Acq{} to define a strategy for efficiently maximizing a \blackbox{} function \f{}. At each ``outer-loop'' iteration (\warn{Figure~\ref{fig:overview_pt1}a}), this strategy is used to choose a set of queries \X{} whose evaluation advances the search process.  
This section reviews related concepts and closes with discussion of the associated inner optimization problem. 
For an in-depth review of BO, we defer to the recent survey \cite{shahriari-ieee16}.

Without loss of generality, we assume BO strategies evaluate \parallelism{}  designs \(\X \in \reals^{\parallelism \times \xdim}\) in parallel so that setting \(\parallelism=1\) recovers purely sequential decision-making. 
We denote available information regarding \f{} as \(\data = \{(\x_{i}, \y_{i})\}_{i=1}^{\ldots}\) and, for notational convenience, assume noiseless observations \(\Y = f(\X)\).
Additionally, we refer to \Acq's parameters (such as an improvement threshold) as \acqParams{} and to \model's parameters as \hypers{}.
%
%
Henceforth, direct reference to these terms will be omitted where possible.

\subheading{Surrogate models}
A surrogate model \model{} provides a probabilistic interpretation of \f{} whereby possible explanations for the function are seen as draws \(\version{\f} \sim p(\f | \data)\). In some cases, this belief is expressed as an explicit ensemble of sample functions \cite{hernandez-nips14,springenberg2016bayesian,wang2017max}. 
More commonly however, \model{} dictates the parameters \(\rparams\) of a (joint) distribution over the function's behavior at a finite set of points \X{}. 
By first tuning the model's (hyper)parameters \(\hypers\) to explain for \(\data\), a belief is formed as \(p(\Y|\X, \data) = p(\Y; \rparams)\) with \(\rparams \gets \model(\X; \hypers)\). Throughout, \(\rparams \gets \model(\X; \hypers)\) is used to denote that belief \(p\)'s parameters \rparams{} are specified by model \model{} evaluated at \X.
A member of this latter category, the Gaussian process prior (GP) is the most widely used surrogate and induces a multivariate normal belief \(\rparams \triangleq (\means,\Cov) \gets \model(\X; \hypers)\) such that \(p(\Y; \rparams) = \gaussian(\Y; \means, \Cov)\) for any finite set \X{} (see \warn{Figure~\ref{fig:overview_pt1}b}).

\subheading{Acquisition functions}
With few exceptions, acquisition functions amount to integrals defined in terms of a belief \(p\) over the unknown outcomes \(\Y = \{\y_{1},\ldots,\y_{\parallelism}\}\) revealed when evaluating a \blackbox{} function \f{} at corresponding input locations \(\X = \{\x_{1},\ldots,\x_{\parallelism}\}\). 
This formulation naturally occurs as part of a Bayesian approach whereby the value of querying \X{} is determined by accounting for the utility provided by possible outcomes \(\sample{\Y} \sim p(\Y|\X, \data)\). 
Denoting the chosen utility function as \acq{}, this paradigm leads to acquisition functions defined as expectations
\begin{align}
\Acq(\X; \data, \acqParams)
	= \E_{\Y}\left[\acq(\Y; \acqParams)\right]
	= \int \acq(\Y; \acqParams) p(\Y|\X, \data{}) d\Y\,.
\label{eq:expected_utility}
\end{align}
A seeming exception to this rule, \emph{non-myopic} acquisition functions assign value by further considering how different realizations of \(\sample{\data}_{\parallelism} \gets \data \cup \{(\x_{i},\sample{\y}_{i})\}_{i=1}^{\parallelism}\) impact our broader understanding of \f{} and usually correspond to more complex, nested integrals. \warn{Figure~\ref{fig:overview_pt1}c} portrays a prototypical acquisition surface and Table~\ref{table:reparameterizations} exemplifies popular, myopic and non-myopic instances of~\eqref{eq:expected_utility}.
\begin{table}
\begin{center}
\newcommand{\YES}{\text{\scriptsize{\textsc{Y}}}}
\newcommand{\NO}{\text{\scriptsize{\textsc{N}}}}

\centering
\setlength\extrarowheight{4pt}
\resizebox{\linewidth}{!}{%
\(\begin{array}{*{5}{c}}
\toprule
	\ {\text{\bfseries Abbr.}}
    & {\text{\bfseries Acquisition Function}\; \Acq{}}
    & {\text{\bfseries Reparameterization}}
    & {\text{\bfseries \MM}}
    \\ \midrule
    
    \small{\EI}
        & \E_{\Y}[\max(\relu(\Y - \alpha))]
        & \E_{\Z}[\max(\relu(\means + \Chol\Z - \alpha))]
        & \YES
    \\
    
    \small{\PI}
        & \E_{\Y}[\max(\heaviside^{-}(\Y - \alpha))]
        & \E_{\Z}[\max(\sigmoid(\tfrac{\means + \Chol\Z - \alpha}{\temperature}))]
        & \YES
    \\
    
    \small{\SR}
        & \E_{\Y}[\max(\Y)]
        & \E_{\Z}[\max(\means + \Chol\Z)]
        & \YES
    \\
    
    \small{\UCB}
        & \E_{\Y}[\max(\means + \sqrt{\nicefrac{\beta\pi}{2}}\abs{\resids})]
        & \E_{\Z}[\max(\means + \sqrt{\nicefrac{\beta\pi}{2}}\abs{\Chol\Z})]
        & \YES
    \\
    
    \small{\ES}
        & -\E_{\Y_{a}}
        [\entropy(
            \E_{\Y_{b}|\Y_{a}}[\heaviside^{+}(\Y_{b} - \max(\Y_{b}))]
        )]
        & -\E_{\Z_{a}}
        [\entropy(
            \E_{\Z_{b}}[
            \softmax(\tfrac{\means_{b|a}+\Chol_{b|a}\Z_{b}}{\temperature})
            ]
        )]
        & \NO
    \\
	
    \small{\KG}
        & \E_{\Y_{a}}[\max(\means_{b} + \Cov_{b,a}\Cov_{a,a}^{-1}(\Y_{a} - \means_{a}))]
        & \E_{\Z_{a}}[\max(\means_{b} + \Cov_{b,a}\Cov_{a,a}^{-1}\Chol_{a}\Z_{a})]
        & \NO
    \\
\bottomrule
\end{array}\)
}
\captionsetup{width=1.0\textwidth}
\vspace{4pt}
\caption{
%
Examples of reparameterizable acquisition functions; the final column indicates whether they belong to the \MM{} family (Section~\ref{sect:myopic_maximal}). 
Glossary: \(\heaviside^{+/-}\) denotes the right-/left-continuous Heaviside step function; \relu{} and \sigmoid{} rectified linear and sigmoid nonlinearities, respectively; \entropy{} the Shannon entropy; \(\alpha\) an improvement threshold; \(\tau\) a temperature parameter; \(\Chol\T{\Chol} \triangleq \Cov\) the Cholesky factor; and, residuals \(\resids \sim \gaussian\left(\zeros, \Cov\right)\). Lastly, non-myopic acquisition function (\ES{} and \KG) are assumed to be defined using a discretization. Terms associated with the query set and discretization are respectively denoted via subscripts \(a\) and \(b\).}
\label{table:reparameterizations}
\end{center}
\vspace{-12pt}
\end{table}

\subheading{Inner optimization problem}
Maximizing acquisition functions plays a crucial role in BO as the process through which abstract machinery (\eg{} model \model{} and acquisition function \Acq) yields concrete actions (\eg{} decisions regarding sets of queries \X). 
Despite its importance however, this inner optimization problem is \temp{often neglected}. 
This lack of emphasis is largely attributable to a greater focus on creating new and improved machinery as well as on applying BO to new types of problems. 
Moreover, elementary examples of BO facilitate \Acq's maximization.
For example, optimizing a single query \(\x \in \reals^{\xdim}\) is usually straightforward when \x{} is low-dimensional and \Acq{} is myopic.

Outside these textbook examples, however, BO's inner optimization problem becomes qualitatively more difficult to solve. 
In virtually all cases, acquisition functions are non-convex (frequently due to the non-convexity of plausible explanations for \f{}).
Accordingly, increases in input dimensionality \xdim{} can be prohibitive to efficient query optimization. In the generalized setting with parallelism \(\parallelism \ge 1\), this issue is exacerbated by the additional scaling in \parallelism. While this combination of non-convexity and (acquisition) dimensionality is problematic, the routine intractability of both non-myopic and parallel acquisition poses a commensurate challenge.

As is generally true of integrals, the majority of acquisition functions are intractable. 
Even Gaussian integrals, which are often preferred because they lead to analytic solutions for certain instances of~\eqref{eq:expected_utility}, are only tractable in a handful of special cases~\cite{genz92numerical,gassmann02,cunningham2011epmgp}.
To circumvent the lack of closed-form solutions, researchers have proposed a wealth of diverse methods. 
Approximation strategies~\cite{cunningham2011epmgp, desautels2014parallelizing, wang2017max}, which replace a quantity of interest with a more readily computable one, work well in practice but may not to converge to the true value.  
In contrast, bespoke solutions~\cite{genz92numerical, ginsbourger2010kriging, chevalier2013fast} provide \mbox{(near-)}analytic expressions but typically do not scale well with dimensionality.\footnote{By \emph{near-analytic}, we refer to cases where an expression contains terms that cannot be computed exactly but for which high-quality solvers exist (\eg{} low-dimensional multivariate normal CDF estimators \cite{genz92numerical,genz2004numerical}).} 
Lastly, \MC methods~\cite{osborne2009gaussian,snoek-nips12a,hennig-jmlr12a} are highly versatile and generally unbiased, but are often perceived as non-differentiable and, therefore, inefficient for purposes of maximizing \Acq{}. 

Regardless of the method however, the (often drastic) increase in cost when evaluating \Acq's proxy acts as a barrier to efficient query optimization, and these costs increase over time as shown in \warn{Figure~1d}.
In an effort to address these problems, we now go inside the outer-loop and focus on efficient methods for maximizing acquisition functions.


\section{Maximizing acquisition functions}
\label{sect:methods}

This section presents the technical contributions of this paper, which can  be broken down into two complementary topics: \inlineItem{1} gradient-based optimization of acquisition functions that are estimated via Monte Carlo integration, and \inlineItem{2} greedy maximization of ``myopic maximal'' acquisition functions. Below, we separately discuss each contribution along with its related literature.


\subsection{Differentiating Monte Carlo acquisitions}
\label{sect:differentiability}
Gradients are one of the most valuable sources of information for optimizing functions. In this section, we detail both the reasons and conditions whereby \MC acquisition functions are differentiable and further show that most well-known examples readily satisfy these criteria (see Table~\ref{table:reparameterizations}).

We assume that \Acq{} is an expectation over a multivariate normal belief \(p(\Y|\X, \data) = \gaussian(\Y; \means, \Cov)\) specified by a GP surrogate such that \((\means, \Cov) \gets \model(\X)\).
More generally, we assume that samples can be generated as \(\sample{\Y} \sim p(\Y|\X, \data)\) to form an unbiased \MC estimator of an acquisition function
\(
\Acq(\X)
	\approx \AcqMC{}(\X)
	\triangleq \tfrac{1}{\nSamples}\sum\nolimits^{\nSamples}_{\sampleIdx=1} \acq(\sample{\Y})
\). 
Given such an estimator, we are interested in verifying whether
\begin{align}
\nabla\Acq(\X)
	\approx \nabla\AcqMC{}(\X)
    \triangleq
    	\tfrac{1}{\nSamples}\sum\nolimits^{\nSamples}_{\sampleIdx=1}
	\nabla\acq(\sample{\Y}),
\label{eq:gradient_mc}
\end{align}
where \(\nabla\acq\) denotes the gradient of utility function \acq{} taken with respect to \X{}. 
The validity of \MC gradient estimator \eqref{eq:gradient_mc} is obscured by the fact that \sample{\Y} depends on \X{} through generative distribution \belief{} and that \(\nabla\AcqMC{}\) is the expectation of \acq's derivative rather than the derivative of its expectation.

Originally referred to as \emph{infinitesimal perturbation analysis}~\cite{cao1985convergence,glasserman1988performance}, the \emph{reparameterization trick}~\cite{kingma2013vae,JimenezRezende2014a} is the process of differentiating through an \MC estimate to its generative distribution \belief's parameters and consists of two components: 
i) reparameterizing samples from \belief{} as draws from a simpler base distribution \(\hat{\belief}\), and 
 ii) interchanging differentiation and integration by taking the expectation over sample path derivatives.
%

\subheading{Reparameterization}
Reparameterization is a way of interpreting samples that makes their differentiability \wrt{} a generative distribution's parameters transparent. Often, samples \(\sample{\Y} \sim p(\Y; \rparams)\) can be re-expressed as a deterministic mapping \(\reparam : \zspace \times \rParams \to \yspace\) of simpler random variates \(\sample{\Z} \sim \hat{p}(\Z)\)~\cite{kingma2013vae, JimenezRezende2014a}. This change of variables helps clarify that, if \acq{} is a differentiable function of \(\Y = \reparam(\Z;\rparams)\), then \(\small{\frac{d\acq}{d\rparams}} = \small{\frac{d\acq}{d\reparam}\frac{d\reparam}{d\rparams}}\) by the chain rule of (functional) derivatives. 

If generative distribution \belief{} is multivariate normal with parameters \(\rparams = (\means, \Cov)\), the corresponding mapping is then \(\reparam(\Z; \rparams) \triangleq \means + \Chol\Z\), where \(\Z \sim \gaussian(\zeros, \identity)\) and \Chol{} is \Cov's Cholesky factor such that \(\Chol\T{\Chol} = \Cov\). Rewriting~\eqref{eq:expected_utility} as a Gaussian integral and reparameterizing, we have
\begin{align}
\Acq(\X) 
	=\int_{\lbounds}^{\ubounds}
		\acq(\Y)
		\gaussian(\Y; \means, \Cov)
		d\Y
	= \int_{\alt{\lbounds}}^{\alt{\ubounds}}
		\acq(\means + \Chol\Z)
		\gaussian(\Z; \zeros, \identity)
		d\Z\,,
\label{eq:expected_utility_mvn}
\end{align}
where each of the \parallelism{} terms \(\alt{c}_i\) in both \(\alt{\lbounds}\) and \(\alt{\ubounds}\) is transformed as \(\alt{c}_{i} = (c_i - \mean_{i} - \sum_{j<i}\chol_{i j}\z_{j})/\chol_{ii}\). The third column of Table~\ref{table:reparameterizations} grounds \eqref{eq:expected_utility_mvn} with several prominent examples. For a given draw \(\sample{\Y} \sim \gaussian(\means, \Cov)\), the sample path derivative of \acq{} \wrt{} \X{} is then
\begin{align}
\nabla\acq(\sample{\Y}) = 
	\frac{d\acq(\sample{\Y})}{d\sample{\Y}}
    \frac{d\sample{\Y}}{d\model(\X)}
    \frac{d\model(\X)}{d\X},
\end{align}
where, by minor abuse of notation, we have substituted in \(\sample{\Y} = \reparam\left(\sample{\Z}; \model(\X)\right)\). 
Reinterpreting \(\Y\) as a function of \(\Z\) therefore sheds light on individual \MC sample's differentiability.

\begin{figure}[t]
\begin{center}
\includegraphics[width=\linewidth]{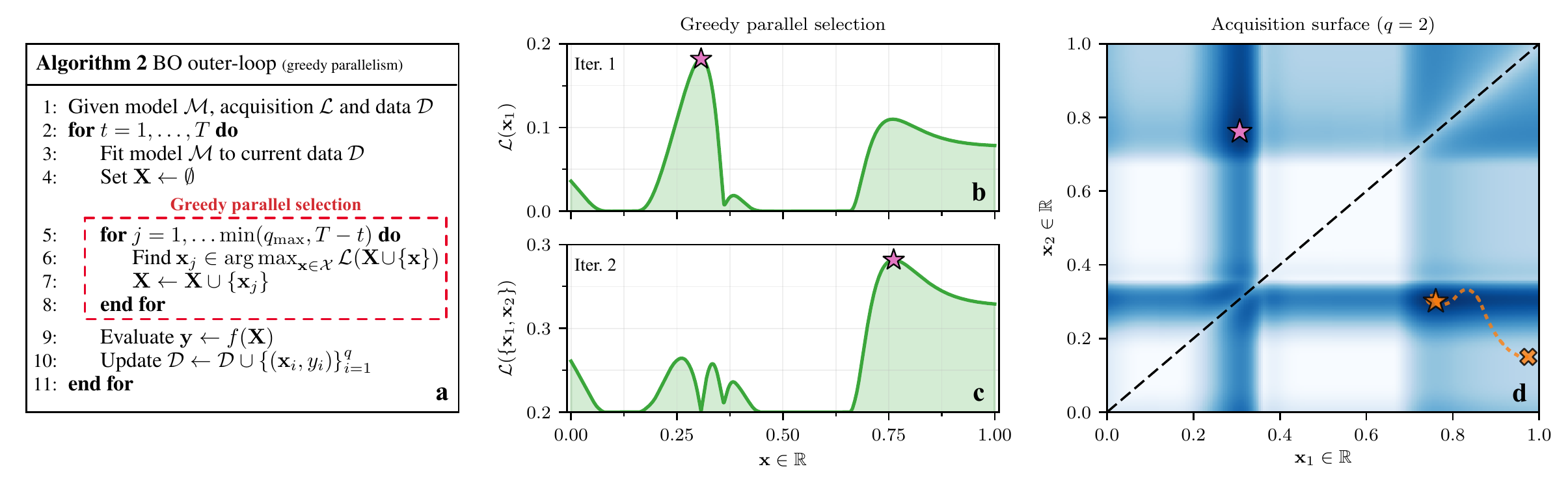}
\vspace{-20pt}
\caption{(a) Pseudo-code for BO outer-loop with greedy parallelism, the inner optimization problem is boxed in red. (b--c) Successive iterations of greedy maximization, starting from the posterior shown in Figure~1b. (d) On the left, greedily selected query `\legendStar'; on the right and from `\legendX' to `\legendStar', trajectory when jointly optimizing parallel queries $\x_{1}$ and $\x_{2}$ via stochastic gradient ascent. Darker colors correspond with larger acquisitions.}
\label{fig:overview_pt2}
\end{center}
\end{figure}

\subheading{Interchangeability}
Since \AcqMC{} is an unbiased \MC estimator consisting of differentiable terms, it is natural to wonder whether the average sample gradient \(\nabla\AcqMC{}\) \eqref{eq:gradient_mc} \temp{follows suit}, \ie{} whether
\begin{align}
\nabla\Acq(\X)
	= 
    	\nabla\E_{\Y}
        \left[\acq(\Y)\right]
    \qeq\E_{\Y}
    	\left[\nabla\acq(\Y)\right]
    \approx 
		\nabla\Acq_{\nSamples}(\X)
\label{eq:gradient_interchange}\,,
\end{align}
where \qeq{} denotes a potential equivalence when interchanging differentiation and expectation. 
\marc{Necessary and sufficient conditions for this interchange are that, as defined under \belief, integrand \acq{} must be continuous and its first derivative \(\acq^\prime\) must a.s.\ exist and be integrable~\citep{cao1985convergence,glasserman1988performance}.}
\citet{wang2016parallel} demonstrated that these conditions are met for a GP with a twice differentiable kernel, provided that the elements in query set \X{} are unique. The authors then use these results to prove that~\eqref{eq:gradient_mc} is an unbiased gradient estimator for the parallel Expected Improvement (\qEI) acquisition function \cite{ginsbourger2010kriging,snoek-nips12a,chevalier2013fast}. In later works, these findings were extended to include parallel versions of the Knowledge Gradient (KG) acquisition function \cite{wu2016parallel,wu2017bayesian}. Figure~\ref{fig:overview_pt2}d (bottom right) visualizes gradient-based optimization of \MC \qEI{} for parallelism \(\parallelism=2\).

\subheading{Extensions}
Rather than focusing on individual examples, our goal is to show differentiability for a broad class of \MC acquisition functions. 
In addition to its conceptual simplicity, one of \MC integration's primary strengths is its generality. This versatility is evident in Table~\ref{table:reparameterizations}, which catalogs (differentiable) reparameterizations for six of the most popular acquisition functions. While some of these forms were previously known (\EI{} and \KG) or follow freely from the above (\SR), others require additional steps. We summarize these steps below and provide full details in Appendix~\ref{appendix:methods}.

In many cases of interest, utility is measured in terms of discrete events. For example, Probability of Improvement \cite{kushner1964new,viana2010surrogate} is the expectation of a binary event \(\event_{\acqSubscript{\PI}}\): ``will a new set of results improve upon a level \level?''
Similarly, Entropy Search \cite{hennig-jmlr12a} contains expectations of categorical events \(\event_{\acqSubscript{\ES}}\): ``which of a set of random variables will be the largest?'' 
Unfortunately, mappings from continuous variables \Y{} to discrete events \event{} are typically discontinuous and, therefore, violate the conditions for \eqref{eq:gradient_interchange}. To overcome this issue, we utilize \emph{concrete} (\underline{con}tinuous to dis\underline{crete}) approximations in place of the original, discontinuous mappings \cite{jang2016categorical,maddison2016concrete}. 

Still within the context of the reparameterization trick, \cite{jang2016categorical,maddison2016concrete} studied the closely related problem of optimizing an expectation \wrt{} a discrete generative distribution's parameters. To do so, the authors propose relaxing the mapping from, \eg{}, uniform to categorical random variables with a continuous approximation so that the (now differentiable) transformed variables closely resemble their discrete counterparts in distribution. Here, we first map from uniform to Gaussian (rather than Gumbel) random variables, but the process is otherwise identical. Concretely, we can approximate \PI's binary event as
\begin{align}
\tilde{\event}_{\acqSubscript{\PI}}(\X; \level, \temperature)
   	= \max\left(\sigmoid\left(\nicefrac{\Y - \alpha}{\temperature}\right)\right)
    \approx \max\left(\heaviside^{-}(\Y - \level)\right),
\end{align}
where \(\heaviside^{-}\) denotes the left-continuous Heaviside step function, \sigmoid{} the sigmoid nonlinearity, and \(\temperature \in [0,\infty]\) acts as a temperature parameter such that the approximation becomes exact as \(\temperature \to 0\). Appendix~\ref{appendix:concrete} further discusses concrete approximations for both \PI{} and \ES{}.

Lastly, the Upper Confidence Bound (\UCB) acquisition function \cite{srinivas10} is typically not portrayed as an expectation, seemingly barring the use of \MC methods. At the same time, the standard definition \(\UCB(\x; \beta) \triangleq \mean + \beta^{\half}\stddev\) bares a striking resemblance to the reparameterization for normal random variables \(\reparam(\z; \mean, \stddev) = \mean + \stddev\z\). By exploiting this insight, it is possible to rewrite this closed-form expression as \(\UCB(\x; \beta) = \int_{\mean}^{\infty} \y\gaussian(\y; \mean, 2\pi\beta\var)d\y\). Formulating \UCB{} as an expectation allows us to naturally parallelize this acquisition function as 
\begin{align}
	\UCB(\X;\beta)
    = \E_{\Y}
    \big[
    	\max(\means + \sqrt{\nicefrac{\beta\pi}{2}}\abs{\resids})
	\big],
\label{eq:parallel_ucb_short}
\end{align}
where \(\abs{\resids} = \abs{\Y - \means}\) denotes the absolute value of \Y{}'s residuals. In contrast with existing parallelizations of \UCB{} \cite{contal2013parallel,desautels2014parallelizing}, Equation~\eqref{eq:parallel_ucb_short} directly generalizes its marginal form and can be efficiently estimated via \MC integration (see Appendix~\ref{appendix:parallel_ucb} for the full derivation).

These extensions further demonstrate how many of the apparent barriers to gradient-based optimization of \MC{} acquisition functions can be overcome by borrowing ideas from new (and old) techniques.

\subsection{Maximizing myopic maximal acquisitions}
\label{sect:myopic_maximal}


This section focuses exclusively on the family of myopic maximal (\MM) acquisition functions: myopic acquisition functions defined as the expected max of a pointwise utility function \uacq{}, \ie{} \(\Acq(\X) = \E_{\Y}[\acq(\Y)] = \E_{\Y}[\max \uacq(\Y)]\). Of the acquisition functions included in Table~\ref{table:reparameterizations}, this family includes \EI{}, \PI, \SR{}, and \UCB{}. \temp{We show that} these functions have special properties that make them particularly amenable to greedy maximization.

Greedy maximization is a popular approach for selecting near-optimal sets of queries \X{} to be evaluated in parallel \cite{azimi2010batch,chen2013near,contal2013parallel,desautels2014parallelizing,shah2015parallel,kathuria2016batched}. This iterative strategy is so named because it always ``greedily'' chooses the query \(\x\) that produces the largest immediate reward. At each step \(\roundIdx = 1, \ldots, \parallelism\), a greedy maximizer treats the \(\roundIdx{-}1\) preceding choices \(\OLD{1}{\X}\) as constants and grows the set by selecting an additional element \(\NEW{0}{\x} \in \argmax_{\x \in \xspace} \Acq(\OLD{1}{\X} \cup \{\x\}; \data)\) from the set of possible queries \xspace{}. \warn{Algorithm~2} in Figure~\ref{fig:overview_pt2} outlines this process's role in BO's outer-loop.



\subheading{Submodularity}
Greedy maximization is often linked to the concept of \emph{submodularity} (\SM). Roughly speaking, a set function \(\Acq\) is \SM{} if its increase in value when adding any new point \NEW{1}{\x} to an existing collection \OLD{1}{\X} is non-increasing in cardinality \(k\) (for a technical overview, see~\cite{bach2013learning}). Greedily maximizing \SM{} functions is guaranteed to produce near-optimal results~\cite{minoux1978accelerated,nemhauser78,krause14}. Specifically, if \Acq{} is a normalized \SM{} function with maximum \(\opt{\Acq}\), then a greedy maximizer will incur no more than \(\tfrac{1}{e}\opt{\Acq}\) regret when attempting to solve for \(\opt{\X} \in \argmax_{\X \in \xspace^{\parallelism}} \Acq(\X)\).

In the context of BO, \SM{} has previously been appealed to when establishing outer-loop regret bounds \cite{srinivas10,contal2013parallel,desautels2014parallelizing}. Such applications of \SM{} utilize this property by relating an idealized BO strategy to greedy maximization of a \SM{} objective (\eg{}, the mutual information between \blackbox{} function \(f\) and observations \data{}). In contrast, we show that the family of \MM{} acquisition functions are inherently \SM, thereby guaranteeing that greedy maximization thereof produces near-optimal choices \X{} at each step of BO's outer-loop.\footnote{An additional technical requirement for \SM{} is that the ground set \xspace{} be finite. Under similar conditions, \SM-based guarantees have been extended to infinite ground sets \cite{srinivas10}, but we have not yet taken these steps.} We begin by removing some unnecessary complexity:

\begin{enumerate}[label=\arabic*.,leftmargin=16pt,rightmargin=8pt]
\item Let \(\sample{f} \sim p(f | \data)\) denote the \(\sampleIdx\)-th possible explanation of \blackbox{} \(f\) given observations \data. By marginalizing out nuisance variables \(f(\xspace \setminus \X)\), \Acq{} can be expressed as an expectation over functions \(\sample{f}\) themselves rather than over potential outcomes \(\sample{\Y} \sim p(\Y|\X, \data)\).

\item 
Belief \(p(f | \data)\) and sample paths \sample{f} depend solely on \data{}. Hence, expected utility \(\Acq(\X; \data) = \E_{f}\left[\acq(f(\X))\right]\) is a weighted sum over a fixed set of functions whose weights are constant.
Since non-negative linear combinations of \SM{} functions are \SM{} \cite{krause14}, \(\Acq(\argdot)\) is \SM{} so long as the same can be said of all functions \(\acq(\sample{f}(\argdot)) = \max \uacq\left(\sample{f}(\argdot)\right)\).
\item
As pointwise functions, \sample{f} and \uacq{} specify the set of values mapped to by \xspace{}. They therefore influences whether we can normalize the utility function such that \(\acq(\emptyset) = 0\), but do not impact \SM{}.   
Appendix \ref{appendix:normalizing_utility} discusses the technical condition of normalization in greater detail. In general however, we require that \(\lboundAcq = \min_{\x \in \xspace}\uacq(\sample{f}(\x))\) is guaranteed to be bounded from below for all functions under the support of \(p(f|\data)\).
\end{enumerate}

Having now eliminated confounding factors, the remaining question is whether \(\max(\cdot)\) is \SM{}.
Let \(\mathcal{V}\) be the set of possible utility values and define \(\max(\emptyset) = \lboundAcq\). 
Then, given sets \(\setA \subseteq \setB \subseteq \mathcal{V}\) and \(\forall v \in \mathcal{V}\), it holds that
\begin{align}
\max(\setA \cup \{v\}) - \max(\setA) \ge \max(\setB \cup \{v\}) - \max(\setB).
\label{eq:max_submodularity}
\end{align}
\temp{\emph{Proof:}} 
We prove the equivalent definition \(\max(\setA) + \max(\setB) \ge \max(\setA~\cup~\setB) + \max(\setA~\cap~\setB)\). Without loss of generality, assume \(\max(\setA~\cup~\setB) = \max(\setA)\). Then, \(\max(\setB) \ge \max(\setA~\cap~\setB)\) since, for any \(\setC \subseteq \setB\), \(\max(\setB) \ge \max(\setC)\).

This result establishes the \MM{} family as a class of \SM{} set functions, providing strong theoretical justification for greedy approaches to solving BO's inner-optimization problem.

\subheading{\warn{Incremental} form}
So far, we have discussed greedy maximizers that select a \roundIdx-th new point \(\NEW{0}{\x}\) by optimizing the joint acquisition
\(
    \Acq(\BOTH{0}{\X}; \data) = \E_{\BOTH{0}{\Y}|\data}\left[\acq(\BOTH{0}{\Y})\right]
\) originally defined in \eqref{eq:expected_utility}. 
A closely related strategy~\cite{ginsbourger2011dealing,snoek-nips12a,contal2013parallel,desautels2014parallelizing} is to formulate the greedy maximizer's objective as (the expectation of) a marginal acquisition function \(\iAcq\). We refer to this category of acquisition functions, which explicitly represent the value of 
\BOTH{0}{\X}
as that of \(\OLD{0}{\X}\) incremented by a marginal quantity, as \emph{\mmform{}}. The most common example of an \mmform{} acquisition function is the iterated expectation
\(
    \E_{\OLD{1}{\Y}|\data}
    \left[
    \iAcq(\NEW{1}{\x}; \data_{\roundIdx})
    \right]
\), where \(\data_{\roundIdx} = \data \cup \{(\x_{i},\y_{i})\}_{i<\roundIdx}\) denotes a fantasy state. Because these integrals are generally intractable, \MC{} integration (Section~\ref{sect:differentiability}) is typically used to estimate their values by averaging over fantasies formed by sampling from \(p(\OLD{1}{\Y}| \OLD{1}{\X}, \data)\).

In practice, approaches based on \mmform{} acquisition functions (such as the mentioned \MC estimator) have several distinct advantages over joint ones. Marginal (myopic) acquisition functions usually admit differentiable, closed-form solutions. The latter property makes them cheap to evaluate, while the former reduces the sample variance of \MC estimators. Moreover, these approaches can better utilize caching since many computationally expensive terms (such as a Cholesky used to generate fantasies) only change between rounds of greedy maximization.

A joint acquisition function \Acq{} can always be expressed as an incremental one by defining \iAcq{} as the expectation of the corresponding utility function \acq's discrete derivative
\begin{align}
\Gain(\NEW{0}{\x}; \OLD{0}{\X}, \data)
  =
    \E_{\BOTH{0}{\Y}|\data}
    \left[
      \gain(\NEW{0}{\y}; \OLD{0}{\Y})
    \right]
  = 
    \Acq(\BOTH{0}{\X}; \data) - \Acq(\OLD{0}{\X}; \data),
\label{eq:growth_expected}
\end{align}
with \(\gain(\NEW{1}{\y}; \OLD{0}{\Y}) = \acq(\BOTH{0}{\Y}) - \acq(\OLD{1}{\Y})\) and \(\Acq(\emptyset; \data) = 0\) so that \(\Acq(\X_{1:q}; , \data) = \sum_{\roundIdx=1}^{\parallelism}\Gain(\NEW{1}{\x}; \OLD{1}{\X}, \data)\). To show why this representation is especially useful for \MM{} acquisition functions, we reuse the notation of \eqref{eq:max_submodularity} to introduce the following straightforward identity
\begin{align}
\max(\setB) - \max(\setA) = \relu\left(\max(\setB \setminus \setA) - \max(\setA)\right).
\label{eq:max_growth}
\end{align}
\emph{Proof:} Since \(\lboundAcq\) is defined as the smallest possible element of either set, the \relu's argument is negative if and only if \setB's maximum is a member of \setA{} (in which case both sides equate to zero). In all other cases, the \relu{} can be eliminated and \(\max(\setB) = \max(\setB \setminus \setA)\) by definition.

Reformulating the \MM{} marginal gain function as \(\gain(\NEW{0}{\y}; \OLD{0}{\Y}) = \relu(\acq(\NEW{0}{\y}) - \acq(\OLD{0}{\Y}))\) now gives the desired result: that the \MM{} family's discrete derivative is the ``improvement'' function. Accordingly, the conditional expectation of \eqref{eq:growth_expected} given fantasy state \(\data_{\roundIdx}\) is the expected improvement of \acq, \ie
\begin{align}
\E_{\NEW{0}{\y}|\data_{\roundIdx}}
    \left[
      \gain(\NEW{0}{\y}; \OLD{0}{\Y})
    \right]
    = \EI_{\acq}\left(\NEW{0}{\x}; \data_{\roundIdx}\right)
    = \int_{\NEW{0}{\Gamma}} \left[\acq(\NEW{0}{\y}) - \acq(\OLD{0}{\Y})\right] p(\NEW{0}{\y}| \NEW{0}{\x}, \data_{\roundIdx}) d\NEW{0}{\y},
\label{eq:growth_expected_ei}
\end{align}
where \(\NEW{1}{\Gamma} \triangleq \{\NEW{1}{\y} : \acq(\NEW{1}{\y}) > \acq(\OLD{1}{\Y})\}\).
Since marginal gain function \(\gain\) primarily acts to lower bound a univariate integral over \NEW{0}{\y}, \eqref{eq:growth_expected_ei} often admits closed-form solutions. This statement is true of all \MM{} acquisition functions considered here, making their \mmform{} forms particularly efficient.

Putting everything together, an \MM{} acquisition function's joint and \mmform{} forms equate as
\(
\Acq(\X_{1:\parallelism}; \data)
  =
  \sum_{\roundIdx=1}^{q} 
    \E_{\OLD{1}{\Y}|\data}
    \left[
        \EI_{\acq}\left(\NEW{0}{\x}; \data_{\roundIdx})\right)
    \right]
\). For the special case of Expected Improvement per se (denoted here as \(\Acq_{\acqSubscript{\EI}}\) to avoid confusion), this expression further simplifies to reveal an exact equivalence whereby
\(
\Acq_{\acqSubscript{\EI}}(\X_{1:\parallelism}; \data)
  = \sum_{\roundIdx=1}^{q}\E_{\OLD{1}{\Y}|\data}
  \left[
    \Acq_{\acqSubscript{\EI}}(\NEW{1}{\x}; \data_{\roundIdx})
  \right]
\). 
Appending~\ref{appendix:results_extended} compares performance when using joint and \mmform{} forms, demonstrating how the latter becomes increasingly beneficial as the dimensionality of the (joint) acquisition function \(\parallelism \times d\) grows.

\newcommand{\ibudget}{inner budget}
\newcommand{\ibudgets}{inner budgets}

\section{Experiments}
\label{sect:experiments}
We assessed the efficacy of gradient-based and submodular strategies for maximizing acquisition function 
in two primary settings: ``synthetic'', where task \f{} was drawn from a known GP prior, and ``black-box'', where \f's nature is unknown to the optimizer. In both cases, we used a GP surrogate with a constant mean and an anisotropic \matern{5}{2} kernel. For black-box tasks, ambiguity regarding the correct function prior was handled via online MAP estimation of the GP's (hyper)parameters. Appendix~\ref{appendix:experiment_details} further details the setup used for synthetic tasks.

We present results averaged over 32 independent trials. Each trial began with three randomly chosen inputs, and competing methods were run from identical starting conditions. While the general notation of the paper has assumed noise-free observations, all experiments were run with Gaussian measurement noise leading to observed values \(\hat{\y} \sim \gaussian(\f(\x), 1\mathrm{e-}3)\).

\subheading{Acquisition functions}
We focused on parallel \MC acquisition functions \AcqMC{}, particularly \EI{} and \UCB{}.
Results using \EI{} are shown here and those using \UCB{} are provided in extended results (Appendix~\ref{appendix:results_extended}). 
To avoid confounding variables when assessing BO performance for different acquisition maximizers, results using the \mmform{} form of \qEI{} discussed in Section~\ref{sect:myopic_maximal} are also reserved for extended results. 

In additional experiments, we observed that optimization of \PI{} and \SR{} behaved like that of \EI{} and \UCB{}, respectively. However, overall performance using these acquisition functions was slightly worse, so further results are not reported here. Across experiments, the \qUCB{} acquisition function introduced in Section~\ref{sect:differentiability} outperformed \qEI{} on all tasks but the Levy function.

Generally speaking, \MC estimators \AcqMC{} come in both deterministic and stochastic varieties. Here, determinism refers to whether or not each of \nSamples{} samples \(\sample{\Y}\) were generated using the same random variates \(\sample{\Z}\) within a given outer-loop iteration (see Section~\ref{sect:differentiability}). 
Together with a decision regarding ``batch-size'' \nSamples{}, this choice reflects a well-known tradeoff of approximation-, estimation-, and optimization-based sources of error when maximizing the true function \Acq{} \cite{bousquet2008tradeoffs}. We explored this tradeoff for each \solver{} and summarize our findings below.

\begin{figure}[t]
\begin{center}
\includegraphics[width=\linewidth]{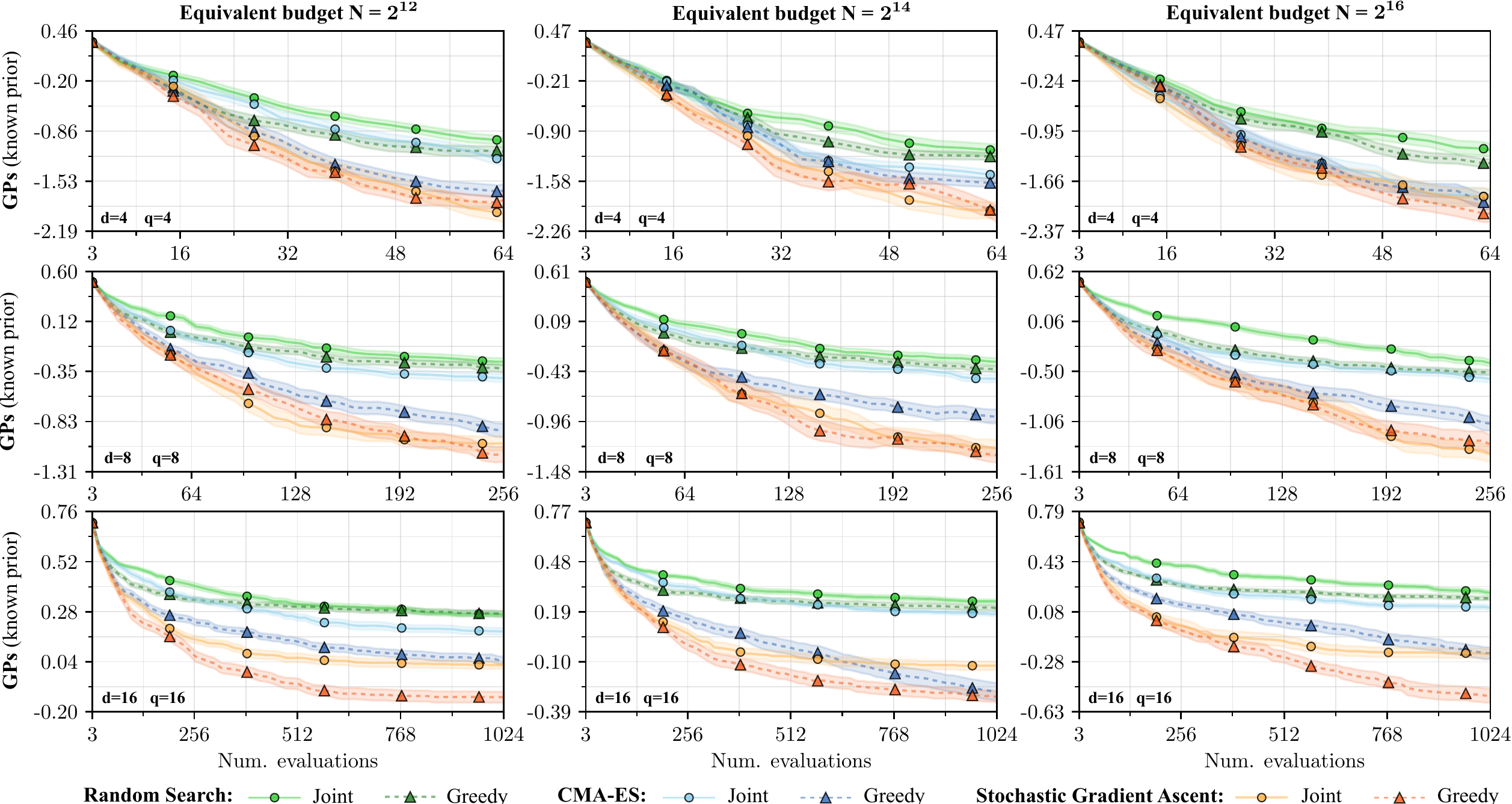}
\caption{Average performance of different acquisition \solvers{} on synthetic tasks from a known prior, given varied runtimes when maximizing Monte Carlo \qEI. Reported values indicate the log of the immediate regret \(\log_{10}\left|f_{\max} - f(\opt{\x})\right|\), where \(\opt{\x}\) denotes the observed maximizer \(\opt{\x} \in \argmax_{\x \in \data} \hat{\y}\).}
\label{fig:results_synthetic_ei}
\end{center}
\end{figure}
\subheading{\Solvers}
We considered a range of (acquisition) \solvers, ultimately settling on stochastic gradient ascent (\adam, \cite{kingma2014adam}), Covariance Matrix Adaptation Evolution Strategy (CMA-ES, \cite{hansen2016cma}) and Random Search (RS, \cite{bergstra12}). Additional information regarding these choices is provided in Appendix~\ref{appendix:experiment_details}.
%
%
For fair comparison, \solvers{} were constrained by CPU runtime. At each outer-loop iteration, an ``\ibudget{}'' was defined as the average time taken to simultaneously evaluate \nEvals{} acquisition values given equivalent conditions. When using greedy parallelism, this budget was split evenly among each of \parallelism{} iterations. To characterize performance as a function of allocated runtime, experiments were run using \ibudgets{} \(\nEvals \in \{2^{12}, 2^{14}, 2^{16}\}\).

For \adam, we used stochastic minibatches consisting of \(\nSamples=128\) samples and an initial learning rate \(\eta=\nicefrac{1}{40}\). To combat non-convexity, gradient ascent was run from a total of 32 (64) starting positions when greedily (jointly) maximizing \AcqMC. Appendix~\ref{appendix:initialization} details the multi-start initialization strategy.
As with the gradient-based approaches, CMA-ES performed better when run using stochastic minibatches \((\nSamples=128)\). Furthermore, reusing the aforementioned initialization strategy to generate CMA-ES's initial population of 64 samples led to additional performance gains.
\nolinebreak 

\begin{figure}[t]
\begin{center}
\includegraphics[width=\linewidth]{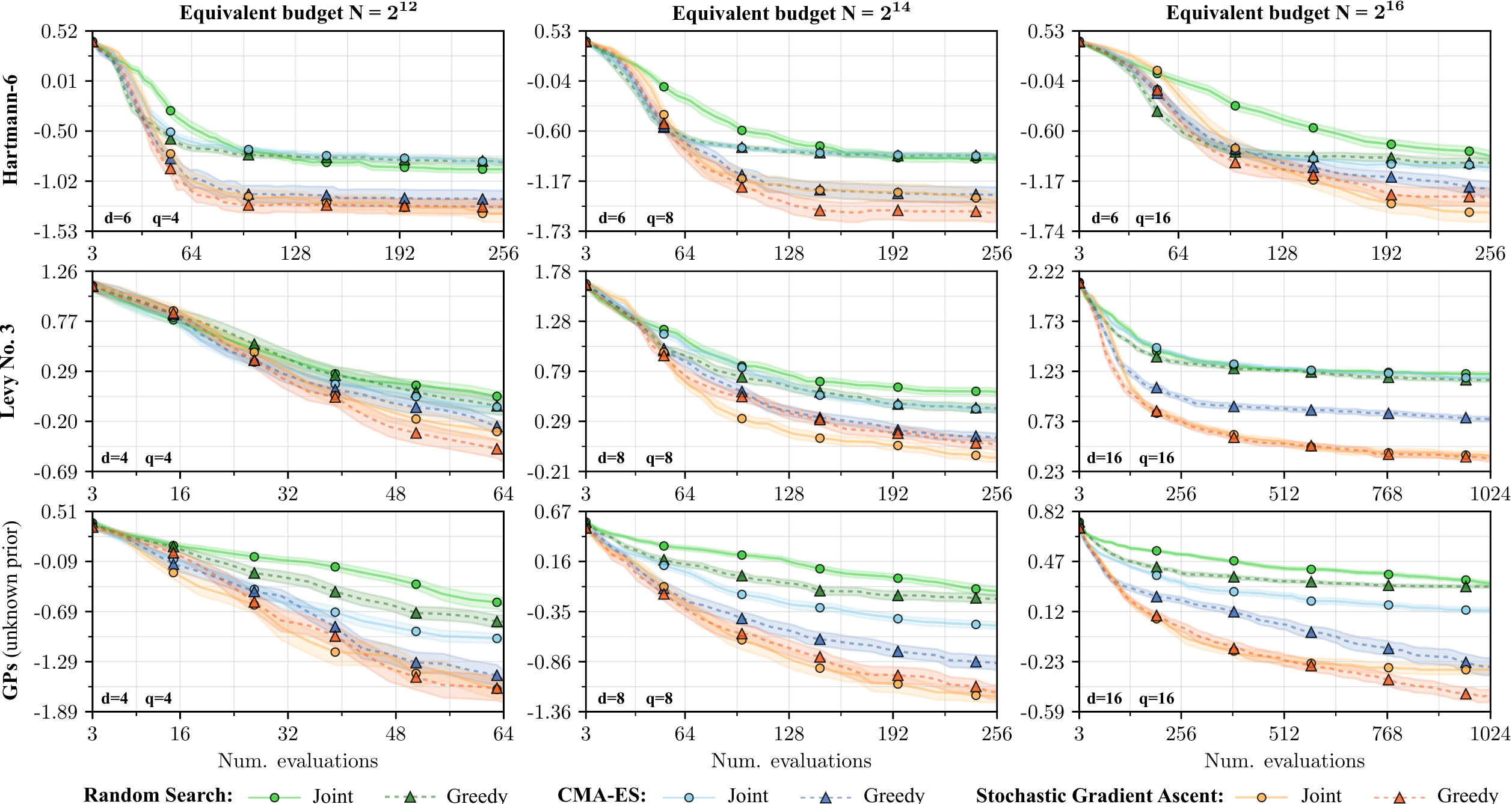}
\caption{Average performance of different acquisition \solvers{} on black-box tasks from an unknown prior, given varied runtimes when maximizing Monte Carlo \qEI. Reported values indicate the log of the immediate regret \(\log_{10}\left|f_{\max} - f(\opt{\x})\right|\), where \(\opt{\x}\) denotes the observed maximizer \(\opt{\x} \in \argmax_{\x \in \data} \hat{\y}\).}
\label{fig:results_blackbox_ei}
\end{center}
\end{figure}

\subheading{Empirical results}
Figures \ref{fig:results_synthetic_ei} and \ref{fig:results_blackbox_ei} present key results regarding BO performance under varying conditions. Both sets of experiments explored an array of input dimensionalities \xdim{} and degrees of parallelism \parallelism{} (shown in the lower left corner of each panel). \Solvers{} are grouped by color, with darker colors denoting use of greedy parallelism; \ibudgets{} are shown in ascending order from left to right. 

Results on synthetic tasks (Figure~\ref{fig:results_synthetic_ei}), provide a clearer picture of the \solvers' impacts on the full BO loop by eliminating the model mismatch. Across all dimensions \xdim{} (rows) and \ibudgets{} \nEvals{} (columns), gradient-based \solvers{} (orange) were consistently superior to both gradient-free (blue) and  na\"ive (green) alternatives. 
Similarly, submodular \solvers{} generally surpassed their joint counterparts. 
However, in lower-dimensional cases where gradients alone suffice to optimize \AcqMC{}, the benefits for coupling gradient-based strategies with near-optima seeking submodular maximization naturally decline. 
Lastly, the benefits of exploiting gradients and submodularity both scaled with increasing acquisition dimensionality \(\parallelism\times\xdim\).

Trends are largely identical for black-box tasks (Figure~\ref{fig:results_blackbox_ei}), and this commonality is most evident for tasks sampled from an unknown GP prior (final row). These runs were identical to ones on synthetic tasks (specifically, the diagonal of Figure~\ref{fig:results_synthetic_ei}) but where knowledge of \f's prior was withheld. Outcomes here clarify the impact of model mismatch, showing how \solvers{} maintain their influence. 
Finally, performance on Hartmann-6 (top row) serves as a clear indicator of the importance for thoroughly solving the inner optimization problem. In these experiments, performance improved despite mounting parallelism due to a corresponding increase in the \ibudget{}.

Overall, these results clearly demonstrate that both gradient-based and submodular approaches to (parallel) query optimization lead to reliable and, often, substantial improvement in outer-loop performance. Furthermore, these gains become more pronounced as the acquisition dimensionality increases. Viewed in isolation, \solvers{} utilizing gradients consistently outperform gradient-free alternatives. Similarly, greedy strategies improve upon their joint counterparts in most cases.


\section{Conclusion}
\label{sect:conclusion}

BO relies upon an array of powerful tools, such as surrogate models and acquisition functions, and all of these tools are sharpened by strong usage practices. 
We extend these practices by demonstrating that Monte Carlo acquisition functions provide unbiased gradient estimates that can be exploited when optimizing them.
%
Furthermore, we show that many of the same acquisition functions form a family of submodular set functions that can be efficiently optimized using greedy maximization.
These insights serve as cornerstones for easy-to-use, general-purpose techniques for practical BO.
Comprehensive empirical evidence concludes that said techniques lead to substantial performance gains in real-world scenarios where queries must be chosen in finite time. 
By tackling the inner optimization problem, these advances directly benefit the theory and practice of Bayesian optimization.

\newpage
\section*{Acknowledgments}
The authors thank David Ginsbourger, Dario Azzimonti and Henry Wynn for initial discussions regarding the submodularity of various integrals.
The support of the EPSRC Centre for Doctoral Training in High Performance Embedded and Distributed Systems (reference EP/L016796/1) is gratefully acknowledged.
This work has partly been supported by the European Research Council (ERC) under the European Union's Horizon 2020 research and innovation programme under grant no. 716721.

\footnotesize
\linespread{1.0}\selectfont
\setlength{\bibsep}{5pt}
\bibliography{references}

\newpage
\normalsize
\appendix

\section{Methods Appendix}
\label{appendix:methods}

\subsection{Concrete approximations}
\label{appendix:concrete}
\begin{figure}[h]
\begin{center}
\includegraphics[width=\linewidth]{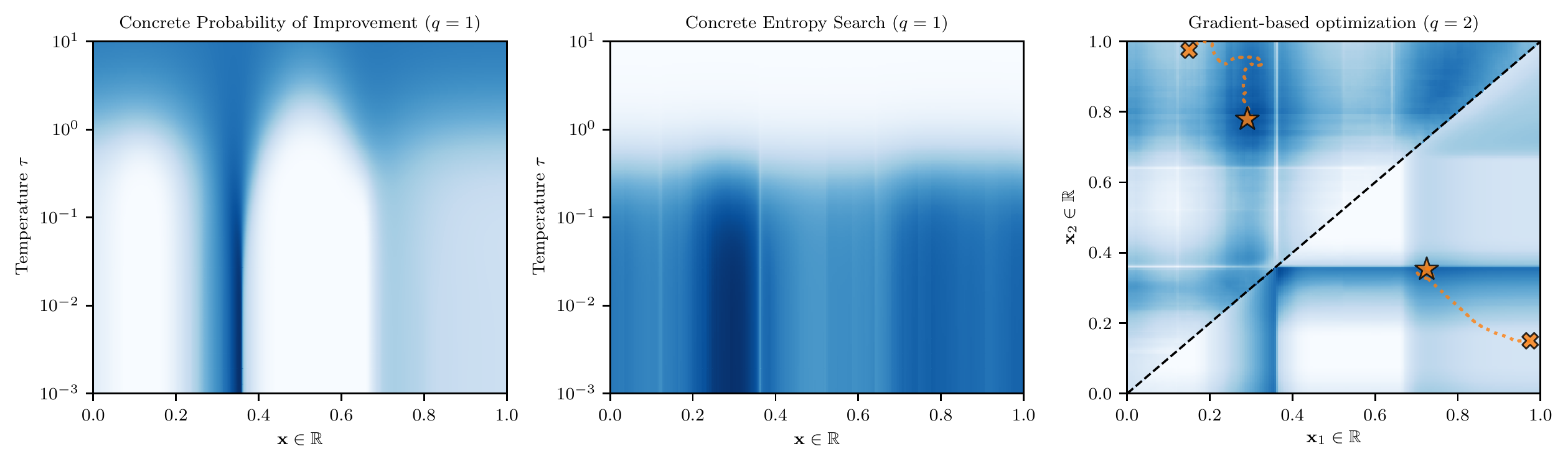}
\caption{Left: Concrete approximation to PI for temperatures \(\temperature \in [1\mathrm{e-}3, 1\mathrm{e-}1]\). Middle: Concrete approximation to ES for temperatures \(\temperature \in [1\mathrm{e-}3, 1\mathrm{e-}1]\). Right: Stochastic gradient ascent trajectories when maximizing concrete approximation to parallel versions of PI (top left) and ES (bottom right), both with a temperature \(\temperature=0.01\).}
\label{fig:concrete_approx}
\end{center}
\end{figure}

As per Section~\ref{sect:differentiability}, utility is sometimes measured in terms of discrete events \(\event \in \espace\). Unfortunately, mappings from continuous values \(\Y \in \yspace\) to the space of discrete events \espace{} are typically discontinuous and, therefore, violate a necessary condition for interchangeability of differentiation and expectation. In the generalized context of differentiating a Monte Carlo integral \wrt{} the parameters of a \emph{discrete} generative distribution, \cite{jang2016categorical,maddison2016concrete} proposed to resolve this issue by introducing a continuous approximation to the aforementioned discontinuous mapping.

As a guiding example, assume that \(\rparams\) is a self-normalized vector of \parallelism{} parameters such that \(\forall \rparam \in \rparams,\, \rparam \ge 0\) and that \(\Z = \T{[\z_{1},\ldots,\z_{\parallelism}]}\) is a corresponding vector of uniform random variables. 
Subsequently, let \(\reparam(\Z; \rparams) = \log(\nicefrac{-\rparams}{\log \Z})\) be defined as random variables \Y's reparameterization. Denoting by \(\opt{\y} = \max(\Y)\), the vector-valued function \(\events : \yspace^{\parallelism}\mapsto \{0, 1\}^{\parallelism}\) defined as
\begin{align}
\events(\Z; \rparams)
	= \T{[\y_{1} \ge \opt{\y},\ldots, \y_{\parallelism} \ge \opt{\y}]}
    = \heaviside^{+}\left(\Y - \opt{\y}\right)
\label{eq:event_onehot}
\end{align}
then reparameterizes a (one-hot encoded) categorical random variable \events{} having distribution \(p(\events; \rparams) = \prod\nolimits_{i=1}^{\parallelism} \rparam_{i}^{\event_{i}}\). Importantly, we can rewrite \eqref{eq:event_onehot} as the zero-temperature limit of as continuous mapping \(\tilde{\events} : \yspace^{\parallelism} \mapsto [0, 1]^{\parallelism}\) defined as 
\begin{align}
\tilde{\events}(\Y; \temperature)
    = \softmax\left(\frac{\Y - \opt{\y}}{\temperature}\right)
    = \softmax\left(\frac{\Y}{\temperature}\right),
\label{eq:event_softmax}
\end{align}
where \(\temperature \in [0, \infty]\) is a temperature parameter. For non-zero temperatures \(\temperature > 0\), we obtain a relaxed version of the original (one-hot encoded) categorical event. Unlike the original however, the relaxed event satisfies the conditions for interchanging differentiation and expectation.

Returning to the case of an acquisition function \eqref{eq:expected_utility} defined over a multivariate normal belief \(p(\Y|\X)\) with parameters \(\rparams = (\means,\Cov)\), random variables \Y{} are instead reparameterized by \(\reparam(\Z; \rparam) = \means + \Chol \operatorname{g}(\Z)\), where \(\operatorname{g}\) denotes, e.g., the Box-Muller transform of uniform \rvs{} \(\Z = \T{[\z_{1},\ldots,\z_{2\parallelism}]}\). 
This particular example demonstrates how Entropy Search's innermost integrand can be relaxed using a \emph{concrete} approximation. Identical logic can be applied to approximate Probability of Improvement's integrand. 

For Monte Carlo versions of both acquisition functions, Figure~\ref{fig:concrete_approx} shows the resulting approximation across a range of temperatures along with gradient-based optimization in the parallel setting \(\parallelism = 2\). Whereas for high temperatures \(\temperature\) the approximations wash out, both converge to the corresponding true function as \(\temperature \to 0\).

\newpage
\subsection{Parallel Upper Confidence Bound (\qUCB{})}
\label{appendix:parallel_ucb}
For convenience, we begin by reproducing \eqref{eq:expected_utility_mvn} as indefinite integrals,
\begin{align*}
\Acq(\X) 
	=\int_{\boldsymbol{-\infty}}^{\boldsymbol{\infty}}
		\acq(\Y)
		\gaussian(\Y; \means, \Cov)
		d\Y
	= \int_{\boldsymbol{-\infty}}^{\boldsymbol{\infty}}
		\acq(\means + \Chol\Z)
		\gaussian(\Z; \zeros, \identity)
		d\Z.
\end{align*}
Working backward through this equation, we derive an exact expression for parallel \UCB{}. To this end, we introduce the definition
\begin{align}
\label{eq:integral_def}
\sqrt{\frac{\pi}{2}}\int_{-\infty}^{\infty}\abs{\stddev\z} \gaussian(\z; 0, 1)d\z 
 	= \sqrt{2\pi}\int_{0}^{\infty} \y \gaussian(\y; 0, \var)d\y = \stddev,
\end{align}
where \(\abs{\cdot}\) denotes the (pointwise) absolute value operator.\footnote{This definition comes directly from the standard integral identity \cite{gradshteyn2014table}: \(\int_{0}^{b}xe^{-q^{2}x^{2}}dx = \frac{1 - e^{-q^{2}b^{2}}}{2q^{2}}\).} Using this fact and given \(\z \sim \gaussian(0, 1)\), let \(\hat{\stddev}^{2} \triangleq (\nicefrac{\beta\pi}{2})\var\) such that \(\E{\abs{\hat{\stddev}\z}} = \beta^{\half}\stddev\). Under this notation, marginal \UCB{} can be expressed as
\begin{align}
\begin{split}
\UCB(\x; \beta)
	&=\mean + \beta^{\half}\stddev\\
	&=\int_{-\infty}^{\infty}
    	\left(\mean + \abs{\hat{\stddev} \z}\right)
    	\gaussian(\z; 0, 1) d\z\\
    &= \int_{-\infty}^{\infty}
    	\left(\mean + \abs{\resid}\right)
    	\gaussian(\resid; 0, \hat{\stddev}^2)d\resid
\end{split}
\label{eq:marginal_ucb}
\end{align}
where \((\mean, \var)\) parameterize a Gaussian belief over \(\y = f(\x{})\) and \(\resid = \y - \mu\) denotes \y{}'s residual. This integral form of \UCB{} is advantageous precisely because it naturally lends itself to the generalized expression
\begin{align}
\begin{split}
\UCB(\X; \beta)
	&= \int_{\boldsymbol{-\infty}}^{\boldsymbol{\infty}}
			\max(\means + \abs{\resids})
			\gaussian(\resids; \zeros, \hat{\Cov})
			d\resids\\
	&= \int_{\boldsymbol{-\infty}}^{\boldsymbol{\infty}}
			\max(\means + \abs{\hat{\Chol}\Z}) 
			\gaussian(\Z; \zeros, \identity)
			d\Z\\
	&\approx 
    	\frac{1}{\nSamples} \sum^{\nSamples}_{\sampleIdx=1} \max(\means + \abs{\hat{\Chol}\sample{\Z}})
        \text{~~for~~}
        \sample{\Z} \sim \gaussian(\zeros, \identity),
\end{split}
\label{eq:parallel_ucb}
\end{align}
where \(\hat{\Chol}\T{\hat{\Chol}} = \hat{\Cov} \triangleq (\nicefrac{\beta\pi}{2})\Cov\). This representation has the requisite property that, for any size \(\alt{\parallelism} \le \parallelism\) subset of \X, the value obtained when marginalizing out the remaining \(\parallelism - \alt{\parallelism}\) terms is its \alt{\parallelism}-UCB value.

Previous methods for parallelizing \UCB{} have approached the problem by imitating a purely sequential strategy \cite{contal2013parallel,desautels2014parallelizing}. Because a fully Bayesian approach to sequential selection generally involves an exponential number of posteriors, these works incorporate various well-chosen heuristics for the purpose of efficiently approximate parallel \UCB{}.\footnote{Due to the stochastic nature of the mean updates, the number of posteriors grows exponentially in \parallelism.} By directly addressing the associated \(\parallelism{}\)-dimensional integral however, Eq.~\eqref{eq:parallel_ucb} avoids the need for such approximations and, instead, unbiasedly estimates the true value.

Finally, the special case of marginal \UCB~\eqref{eq:marginal_ucb} can be further simplified as
\begin{align}
\label{eq:mUCB_simple}
\UCB(\x; \beta)
    =\mean + 2\int_{0}^{\infty}\hat{\stddev} \z \gaussian(\z; 0, 1)dz
    =\int_{\mu}^{\infty} \y \gaussian(\y; \mu, 2\pi\beta\var) d\y,
\end{align}
revealing an intuitive form --- namely, the expectation of a Gaussian random variable (with rescaled covariance) above its mean.
\pagebreak

\subsection{Normalizing utility functions}
\label{appendix:normalizing_utility}

An additional requirement when proving the near-optimality of greedy maximization for a \SM{} function \Acq{} is that \Acq{} be a normalized set function such that \(\Acq(\emptyset) = 0\).
As in Section~\ref{sect:myopic_maximal}, let \lboundAcq{} be defined as the smallest possible utility value given a utility function \acq{} defined over a ground set \version{f} indexed by \xspace{}. Because the \(\max\) is additive such that \(\max(\Y - \lboundAcq) = \max(\Y) - \lboundAcq\), normalization is only necessary when establishing regret bounds and simply requires lower bounding \(\lboundAcq{} > -\infty\). This task is is facilitated by the fact that \lboundAcq{} pertains to the outputs of \acq{} rather than to (a belief over) black-box \f{}.
\temp{Addressing the matter by case, we have:}
\begin{enumerate}[label=\alph*.,leftmargin=16pt]
\item Expected Improvement: For a given threshold \level{}, let improvement be defined (pointwise) as \(\relu(\Y - \level) = \max(0, \Y - \level)\). \EI's integrand is then the largest improvement \(\acq_{\acqSubscript{\EI}}(\Y; \level) = \max(\relu(\Y - \level))\).\footnote{\(\acq_{\acqSubscript{\EI}}\) is often written as the improvement of \(\max(\Y)\); however, these two forms are equivalent.} Applying the rectifier prior to the \(\max\) defines \(\acq_{\acqSubscript{\EI}}\) as a normalized, submodular function.
\item Probability of Improvement: \PI's integrand is defined as \(\acq_{\acqSubscript{\PI}}(\Y, \alpha) = \max(\heaviside^{-}(\Y - \alpha))\), where \(\heaviside^{-}\) denotes the left-continuous Heaviside step function. Seeing as the Heaviside maps \(\yspace \mapsto \{0,1\}\), \(\acq_{\acqSubscript{\PI}}\) is already normalized.
\item Simple Regret: The submodularity of Simple Regret was previously discussed in \cite{azimi2010batch}, under the assumption \(\lboundAcq = 0\). More generally, normalizing \(\acq_{\acqSubscript{\SR}}\) requires bounding \f's infimum under \(p\). Technical challenges for doing so make submodular maximization of SR the hardest to justify.
\item Upper Confidence Bound: As per \eqref{eq:parallel_ucb_short}, define \UCB's integrand as the maximum over \Y's expectation incremented by non-negative terms. 
By definition then, \(\acq_{\acqSubscript{\UCB}}\) is lower bounded by the predictive mean and can therefore be normalized as \(\bar{\acq}_{\acqSubscript{\UCB}} = \max(\means + \abs{\resids} - \lboundAcq)\), provided that \(\lboundAcq = \min_{\x \in \xspace}\mean(\x)\) is finite. 
For a zero-mean GP with a twice differentiable kernel, this condition is guaranteed for bounded functions \f{}.
\end{enumerate}

\subsection{Expected Improvement's \mmform{} form}

For the special case of \(\Acq_{\acqSubscript{\EI}}\), the expected improvement of improvement integrand \(\acq_{\acqSubscript{\EI}}\) simplifies as:
\begin{align}
\begin{split}
\EI_{\acq_{\acqSubscript{\EI}}}(\NEW{1}{\x}, \data_{\roundIdx})
    &= 
        \E_{\NEW{1}{\y}}\left[
        \relu\left(
          \relu(\NEW{1}{\y} - \level) - \max \relu(\OLD{1}{\Y} - \level)
        \right)\right]\\
    &= 
        \E_{\NEW{1}{\y}}\left[
        \relu\left(
          \max(\level, \NEW{1}{\y}) - \max(\level, \max \OLD{1}{\Y})
        \right)\right]\\
    &= 
        \E_{\NEW{1}{\y}}\left[
        \relu\left(
          \NEW{1}{\y} - \max(\level, \max \OLD{1}{\Y})
        \right)\right]\\
    &= 
        \Acq_{\acqSubscript{\EI}}(\NEW{1}{\x}; \data_{\roundIdx}),
\end{split}
\label{eq:\mmform{}_ei}
\end{align}
where \(\alpha = \max(\{y : \forall (\x, \y) \in \data\})\) denotes the initial improvement threshold.

\section{Experiments Appendix}
\label{appendix:experiments}

\subsection{Experiment Details}
\label{appendix:experiment_details}

\subheading{Synthetic tasks} To eliminate model error, experiments were first run on synthetic tasks drawn from a known prior. For a GP with a continuous, stationary kernel, approximate draws \f{} can be constructed via a weighted sum of basis functions sampled from the corresponding Fourier dual \cite{bochner1959lectures, rasmussen-book06a, rahimi2008random, hernandez-nips14}. For a \matern{$\nu$}{2} kernel with anisotropic lengthscales \(\Precis^{-1}\), the associated spectral density is the multivariate \(\mathnormal{t}\)-distribution \(t_{\nu}(\zeros, \Precis^{-1})\) with \(\nu\) degrees of freedom \cite{kotz2004multivariate}. For our experiments, we set \(\Precis = (\nicefrac{\xdim}{16})\,\identity \) and approximated \f{} using \(2^{14}\) basis functions, \temp{resulting in} tasks that were sufficiently challenging and closely resembled exact draws from the prior.

\subheading{Maximizers} In additional to findings reported in the text, we compared several gradient-based approaches (incl. L-BFGS-B and Polyak averaging~\cite{wang2016parallel}) and found that \adam{} consistently delivered superior performance. CMA-ES was included after repeatedly outperforming the rival black-box method \direct{} \cite{jones1993lipschitzian}. RS was chosen as a na\"ive acquisition \solver{} after Successive Halving (SH, \cite{karnin2013almost,jamieson2016non}) failed to yield significant improvement.
For consistency when identifying the best proposed query set(s), both RS and SH used deterministic estimators \AcqMC{}. Whereas RS was run with a constant batch-size \(\nSamples=1024\), SH started small and iteratively increased \nSamples{} to refine estimated acquisition values for promising candidates using a cumulative moving average and cached posteriors.

\subsection{Multi-start initialization}
\label{appendix:initialization}
As noted in \cite{wang2016parallel}, gradient-based query optimization strategies are often sensitive to the choice of starting positions. This \temp{sensitivity} naturally occurs for two primary reasons. 
First, acquisition functions \Acq{} are consistently non-convex. As a result, it is easy for members of query set \(\X \subseteq \xspace\) to get stuck in local regions of the space. 
Second, acquisition surfaces are frequently patterned by (large) plateaus offering little expected utility. Such plateaus typically emerge when corresponding regions of \xspace{} are thought to be inferior and are therefore excluded from the search process.

To combat this issue, we appeal to the submodularity of \Acq{} (see Section~\ref{sect:myopic_maximal}). Assuming \Acq{} is submodular, then acquisition values exhibit diminishing returns \wrt{} the degree of parallelism \parallelism{}. As a result, the marginal value for querying a single point \(\x \in \xspace\) upper bounds its potential contribution to any query set \X{} s.t. \(\x \in \X\). Moreover, marginal acquisition functions \(\Acq(\x)\) are substantially cheaper to compute that parallel ones (see Figure~\ref{fig:overview_pt1}d). Accordingly, we can initialize query sets \X{} by sampling from \(\Acq(\x)\). By doing so we \temp{gracefully avoid initializing points in excluded regions}, mitigating the impact of acquisition plateaus.

In our experiments we observed consistent performance gains when using this strategy in conjunction with most query optimizers. To accommodate runtime constraints, the initialization process was run for the first tenth of the allocated time.

Lastly, when greedily maximizing \Acq{} (equiv. in parallel asynchronous cases), ``pending'' queries were handled by fantasizing observations at their predictive mean. Conditioning on the expected value reduces uncertainty in the vicinity of the corresponding design points and, in turn, promotes diversity within individual query sets \cite{desautels2014parallelizing}. To the extent that this additional step helped in our experiments, the change in performance was rather modest.

\subsection{Extended Results}
\label{appendix:results_extended}

Additional results for both \qUCB{} (Section~\ref{sect:differentiability}) and \mmform{} form \qEI{} (Section~\ref{sect:myopic_maximal}) are shown here. These experiments were run under identical conditions to those in Section~\ref{sect:experiments}. 

\subheading{Parallel UCB} We set confidence parameter \(\beta = 2\). Except for on the Levy benchmark, \qUCB{} outperformed \qEI{}, and this result also held for both Branin-Hoo and Hartmann-3 (not shown).

\subheading{Incremental q-EI} We tested performance using \(\nSamples \in \{16, 32, 64, 128\}\) states. At the end of the first round of greedy selection \(k=1\), \nSamples{} outcomes \(\y^{(i)}_{1} \sim p(\y_{1} | \x_{1}, \data)\) were fantasized, producing \nSamples{} distinct fantasy states \(\data^{(i)}_{1} = \data \cup \{(\x_{1}, \y^{(i)}_{1})\}\). At all other steps \(k \in [2,\parallelism]\), a single outcome was fantasize for each state such that the number of states remained constant. Additionally, fantasized outcomes were never resampled.

Figure~\ref{fig:results_synthetic_ei_\mmform{}} compares results obtained when greedily maximizing \mmform{} \qEI{} (with \(\nSamples=16\) to those obtained when greedily maximizing joint \qEI{} (as discussed in Section~\ref{sect:experiments}). In contrast with the larger body of results, CMA-ES combined with \mmform{}
\qEI{} outperformed gradient-based optimization for higher dimensional acquisition surfaces.


\begin{figure}[t]
\begin{center}
\includegraphics[width=\linewidth]{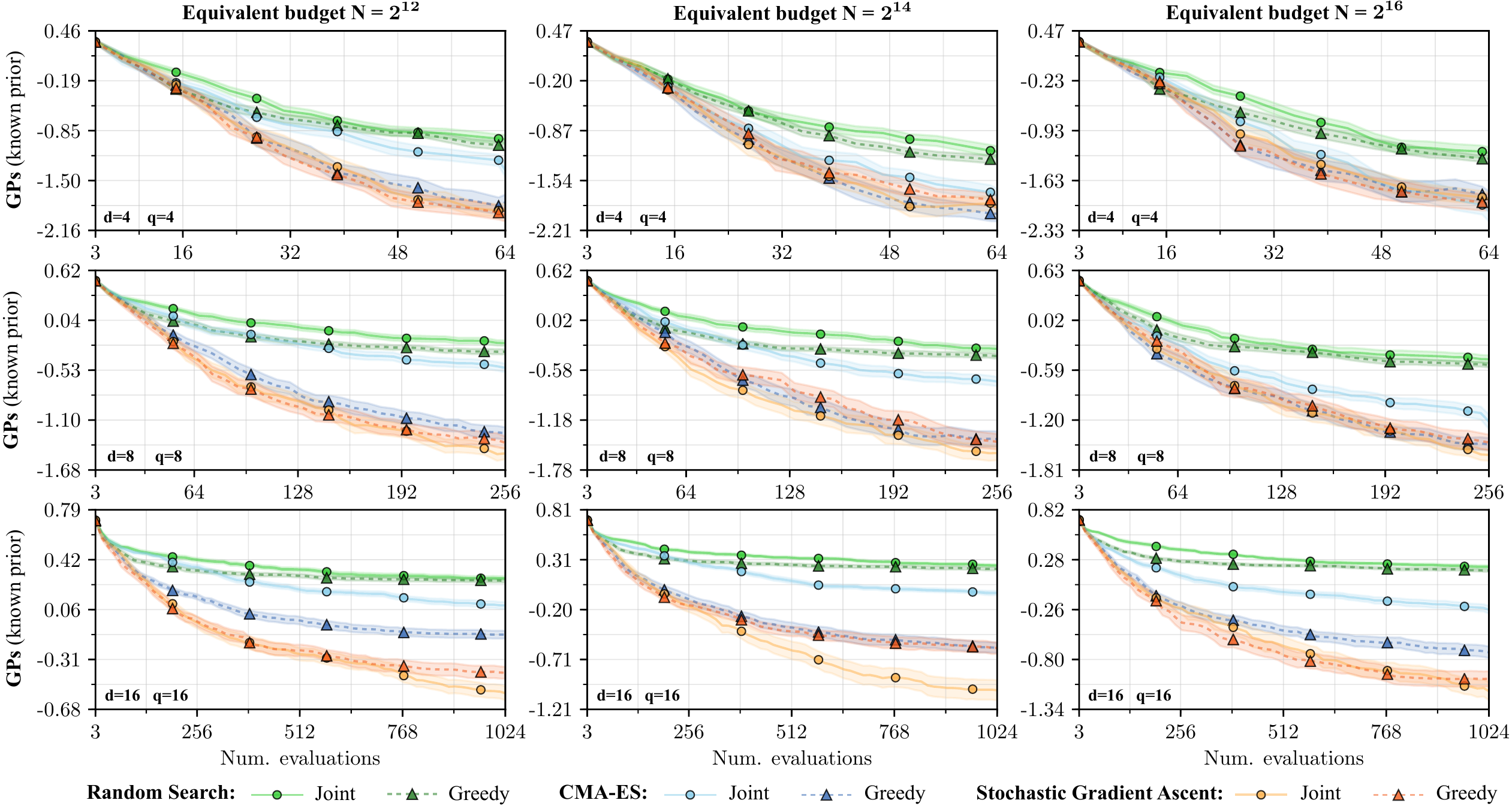}
\caption{Average performance of different acquisition \solvers{} on synthetic tasks from a known prior, given varied runtimes when maximizing Monte Carlo \qUCB. Reported values indicate the log of the immediate regret \(\log_{10}\left|f_{\max} - f(\opt{\x})\right|\), where \(\opt{\x}\) denotes the observed maximizer \(\opt{\x} \in \argmax_{\x \in \data} \hat{\y}\).}
\label{fig:results_synthetic_ucb}
\end{center}
\end{figure}

\begin{figure}
\begin{center}
\includegraphics[width=\linewidth]{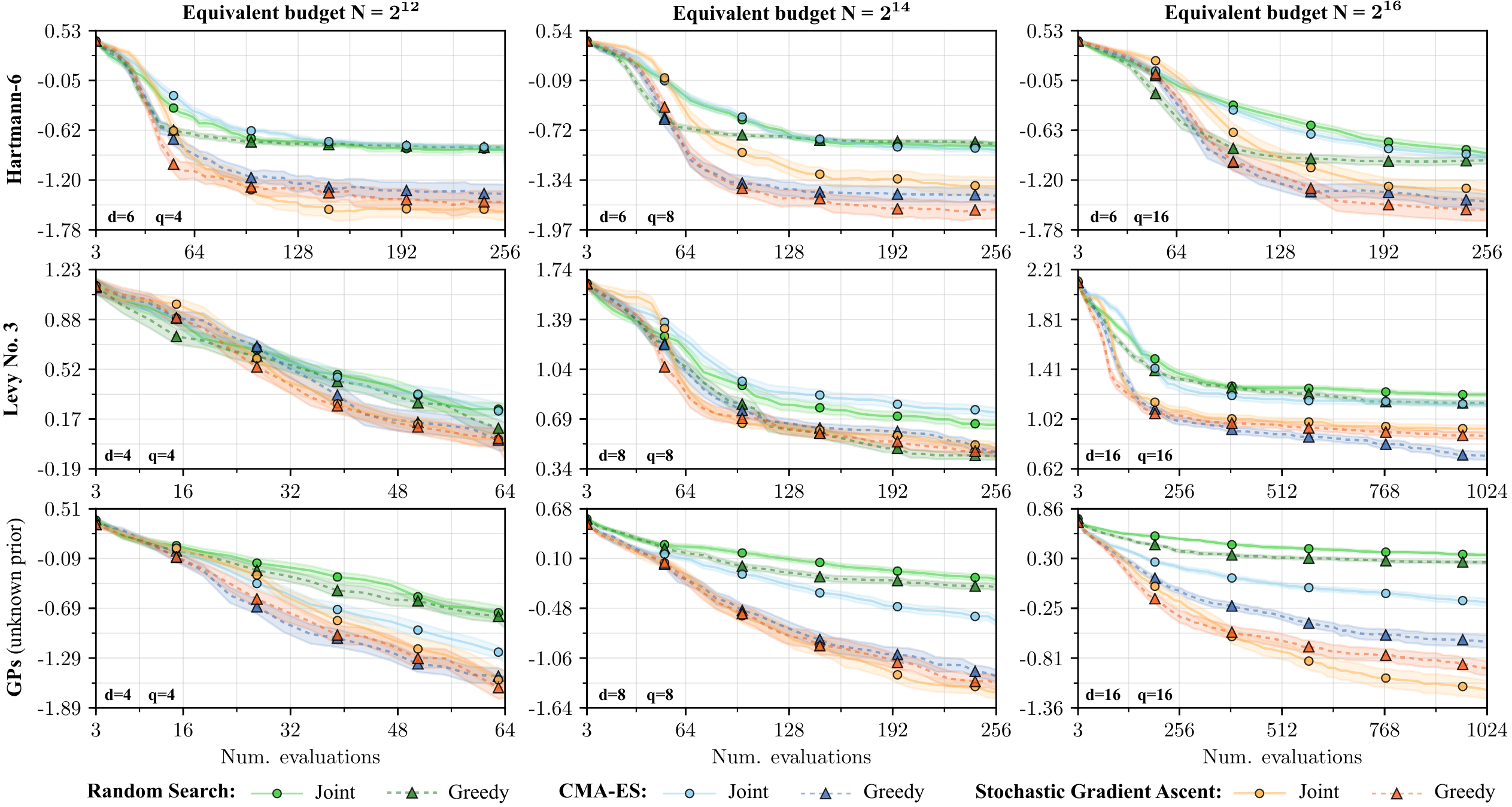}
\caption{Average performance of different acquisition \solvers{} on black-box tasks from an unknown prior, given varied runtimes when maximizing Monte Carlo \qUCB. Reported values indicate the log of the immediate regret \(\log_{10}\left|f_{\max} - f(\opt{\x})\right|\), where \(\opt{\x}\) denotes the observed maximizer \(\opt{\x} \in \argmax_{\x \in \data} \hat{\y}\).}
\label{fig:results_blackbox_ucb}
\end{center}
\end{figure}

\begin{figure}[t]
\begin{center}
\includegraphics[width=\linewidth]{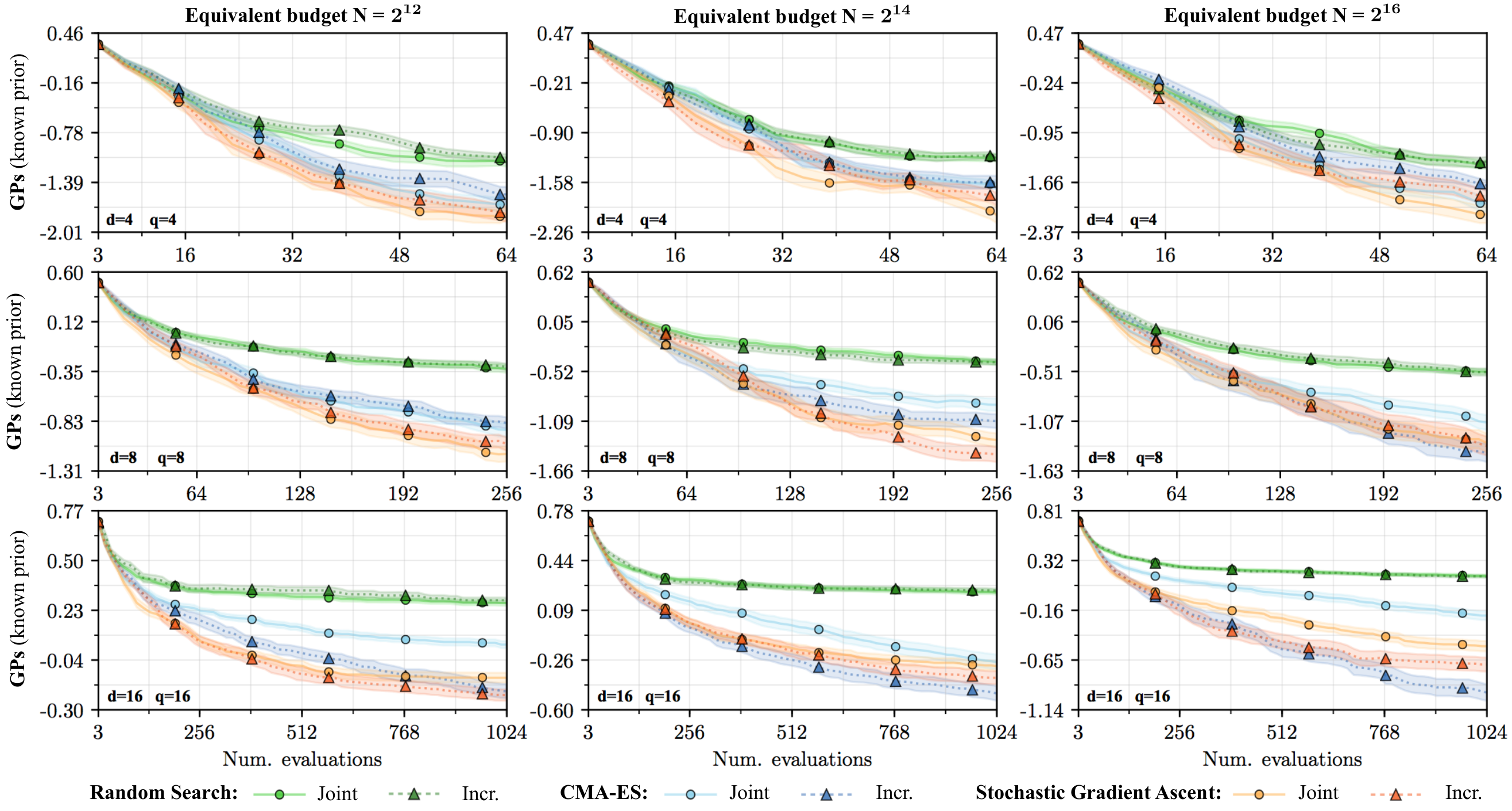}
\caption{Average performance when greedily maximizing joint vs. \mmform{} forms of \qEI{}.}
\label{fig:results_synthetic_ei_\mmform{}}
\end{center}
\end{figure}

\end{document}